\documentclass{article}

\usepackage[preprint, nonatbib]{neurips_2022}

\usepackage[utf8]{inputenc} % allow utf-8 input
\usepackage[T1]{fontenc}    % use 8-bit T1 fonts
\usepackage{hyperref}       % hyperlinks
\usepackage{url}            % simple URL typesetting
\usepackage{booktabs}       % professional-quality tables
\usepackage{amsfonts}       % blackboard math symbols
\usepackage{nicefrac}       % compact symbols for 1/2, etc.
\usepackage{microtype}      % microtypography
\usepackage{xcolor}         % colors
\usepackage{csvsimple}
\usepackage{amsmath,amssymb,amstext} % Lots of math symbols and environments
\usepackage{graphicx}

\title{GHN-QAT: Training Graph Hypernetworks to Predict Quantization-Robust Parameters of Unseen Limited Precision Neural Networks}

\author{%
  Stone Yun\\
  Vision and Image Processing Research Group, University of Waterloo\\
  Waterloo Artificial Intelligence Institute, Waterloo, Canada\\
  \texttt{s22yun@uwaterloo.ca} \\
  \And
  Alexander Wong\\
  Vision and Image Processing Research Group, University of Waterloo\\
  Waterloo Artificial Intelligence Institute, Waterloo, Canada\\
  DarwinAI Corp., Waterloo, Canada\\
  \texttt{a28wong@uwaterloo.ca} \\
}

\begin{document}
\maketitle

%%%%%%%%% ABSTRACT
\begin{abstract}
   Graph Hypernetworks (GHN) can predict the parameters of varying unseen CNN architectures with surprisingly good accuracy at a fraction of the cost of iterative optimization. Following these successes, preliminary research has explored the use of GHNs to predict quantization-robust parameters for 8-bit and 4-bit quantized CNNs. However, this early work leveraged full-precision float32 training and only quantized for testing. We explore the impact of quantization-aware training and/or other quantization-based training strategies on quantized robustness and performance of GHN predicted parameters for low-precision CNNs.  We show that quantization-aware training can significantly improve quantized accuracy for GHN predicted parameters of 4-bit quantized CNNs and even lead to greater-than-random accuracy for 2-bit quantized CNNs. These promising results open the door for future explorations such as investigating the use of GHN predicted parameters as initialization for further quantized training of individual CNNs, further exploration of "extreme bitwidth" quantization, and mixed precision quantization schemes.
\end{abstract}

%%%%%%%%% BODY TEXT
\vspace{-0.2in}

\begin{figure}

\centerline{\includegraphics[width=13cm]{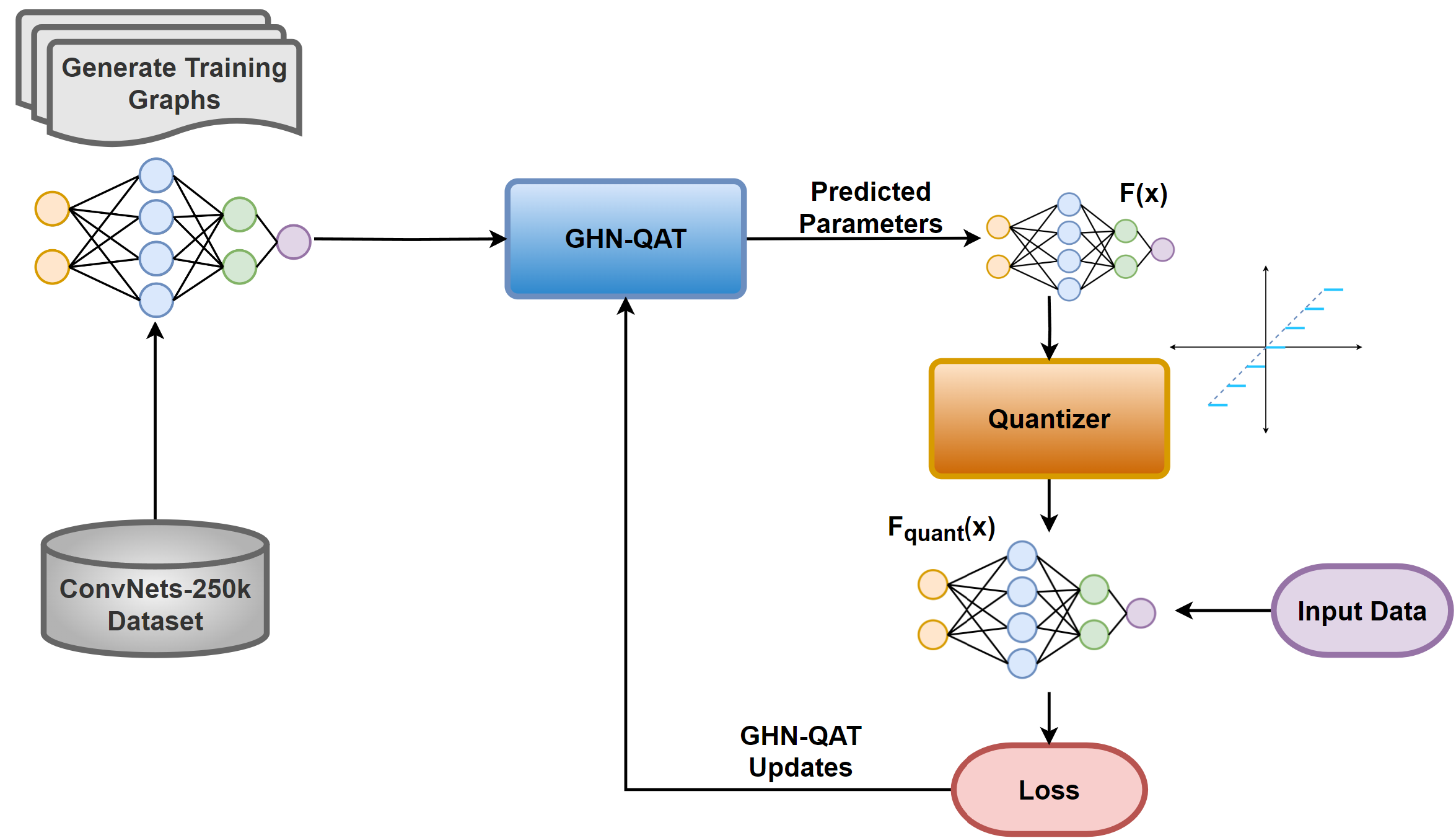}}
\caption[ghnq\_training]{GHN-QAT finetuned on ConvNets-250K. We generate a large number of CNN graphs which are then quantized to target bitwidth for training. Once trained, GHN-QAT can predict robust parameters for unseen CNNs.}
\label{fig:GHNQ_train}
\end{figure}

\section{Introduction}
\label{sec:intro}
\vspace{-0.05in}

Low-bit neural networks that use limited precision quantization for inference~\cite{Nagel_DFQ, TFQuant} can make state-of-the-art models much more practical. However, perturbations induced by quantization of weights and activations can change DNN behaviour in non-trivial ways. In some cases, state-of-the-art performance can have significant degradation after quantization of weights and activations~\cite{MobileNetsQuantizePoorly}. So how can we find high-performant, quantization robust CNN parameters?

Recent works by~\cite{PPUDA}~and~\cite{GHN1} have shown remarkable performance using Graph Hypernetworks (GHN) to predict all trainable parameters of unseen DNNs in a \textit{single forward pass}. Preliminary research in \cite{ghnq_eiw} has explored the use of GHNs to predict quantization-robust parameters for 8-bit and 4-bit quantized CNNs. However, this work trained GHNs to predict parameters for full-precision float32 candidate CNNs and only quantized them for testing. Building on this, we explore quantization-specific training and find that quantization-aware training of GHNs (which we refer to as GHN-QAT) can significantly improve quantized accuracy for GHN predicted parameters of 4-bit quantized CNNs and even achieve greater-than-random accuracy for 2-bit CNNs. More specifically, we simulated quantization (SimQuant) in sampled CNNs such that GHN-QAT adapts to the quantization errors induced by quantizing GHN-predicted models (see Fig.~\ref{fig:GHNQ_train}). By finetuning GHNs on a mobile-friendly, \textbf{quantized} CNN architecture space, GHNs learn representations specifically for efficient quantized CNNs.

\vspace{-0.1in}
\section{Experiment}
\label{sec:experiment}
\vspace{-0.05in}

We first investigate SimQuant based quantization training (commonly referred to as quantization-aware training/QAT) on a target design space for limited precision quantization. We evaluate a few low bit-width settings (W4/A4, W4/A8, W2/A2 where W indicates weight bitwidth and A indicates activation bitwidth). Using SimQuant for W2/A2 proved to be unstable and we found that modelling quantization as uniform noise (NoiseQuant) led to much better results. The reported W2/A2 results are from training with NoiseQuant where the sampling distribution is computed based on 2-bit precision. In all cases, GHN-QAT training is precision/bitwidth-specific. Encoding bitwidth into the CNN graph could potentially remove the need for bit-width specific finetuning. We finetuned a CIFAR-10, DeepNets-1M pretrained GHN-2 model obtained from~\cite{PPUDA_Github} on the ConvNets-250K graph dataset from~\cite{ghnq_eiw}. We use tensorwise, asymmetric, uniform quantization throughout. Figure~\ref{fig:GHNQ_train} shows how GHN-QAT is finetuned to predict efficient, quantization-robust CNNs.

GHN-QAT was finetuned on ConvNets-250K for 100 epochs using CIFAR-10~\cite{cifar10}. We follow a testing procedure similar to~\cite{PPUDA},~\cite{ghnq_eiw} and evaluate GHN-QAT by comparing the mean CIFAR-10 test accuracy at the stated target precisions. Table~\ref{tab:results} shows the top-1 and top-5 accuracy results on different testing splits. To establish the benefits of QAT, we also include results from~\cite{ghnq_eiw} where the authors used full-precision float32 training and only quantized CNNs for testing. In~\cite{ghnq_eiw}, W2/A2 degraded to random accuracy. We use weight-tiling and normalization as described in~\cite{PPUDA} and use $s^{(max)}=10$ for max shortest path of virtual edges.

\begin{table*}[h]
    \caption{Testing GHN-QAT on unseen quantized networks. CIFAR-10 test accuracy of quantized CNNs following bitwidth-specific QAT. Presented as (Mean$\%\pm$standard error of mean; Max$\%$). Rows 1 and 2 show results from~\cite{ghnq_eiw}. \textbf{ID} indicates in-distribution graphs sampled the same way as training set. \textbf{OOD} are out-of-distribution graphs with characteristics very different from those sampled in training. BN-Free networks have no BatchNorm layers. Wide/Deep indicate much wider/deeper nets than those seen in training.}
    
    \centering
    \begin{tabular}{c| c| ccc}
    \toprule
     \textbf{Top-1 Accuracy} & \textbf{ID} & \multicolumn{3}{c}{\textbf{OOD}}\\% \vline\\
     \textbf{by Bitwidth} & Test & Deep &  Wide &  BN-Free \\
     \midrule
      W4/A4~\cite{ghnq_eiw} & $37.2\pm0.3; 52.6$ & $30.7\pm0.5; 50.2$ & $34.7\pm0.5; 50.7$ & $21.5\pm0.8; 36.7$\\
      W4/A8~\cite{ghnq_eiw} & $47.4\pm 0.4; 63.9$ & $39.0\pm 0.7; 62.9$ & $43.8\pm 0.6; 63.3$ & $25.7\pm 0.9; 43.7$\\
      W4/A4 & $52.5\pm0.4; 65.7$ & $50.3\pm0.6; 63.7$ & $50.6\pm0.6; 64.1$ & $24.8\pm0.7; 37.2$ \\
      W4/A8 & $60.2\pm0.4; 75.3$ & $56.5\pm0.7; 72.5$ & $58.1\pm0.7; 73.5$ & $30.7\pm1.0; 49.2$ \\
      W2/A2 & $25.6\pm0.2; 36.3$ & $23.0\pm0.3; 34.4$ & $24.6\pm0.4; 34.8$ & $10.6\pm0.1; 14.2$\\
    
    \toprule
     \textbf{Top-5 Accuracy} & \textbf{ID} & \multicolumn{3}{c}{\textbf{OOD}}\\% \vline\\
     \textbf{by Bitwidth} & Test & Deep &  Wide &  BN-Free \\
     \midrule
      W4/A4~\cite{ghnq_eiw} & $83.7\pm0.2; 91.7$ & $78.6\pm0.4; 91.0$ & $82.1\pm0.3; 91.4$ & $69.3\pm1.1; 87.0$\\
      W4/A8~\cite{ghnq_eiw} & $89.8\pm 0.2; 95.9$ & $85.2\pm 0.4; 95.8$ & $88.0\pm 0.3; 95.5$ & $74.6\pm 1.2; 89.8$\\
      W4/A4 & $91.9\pm0.1; 96.1$ & $90.5\pm0.3; 95.9$ & $91.3\pm0.2; 95.7$ & $76.5\pm0.9; 89.2$ \\
      W4/A8 & $94.5\pm0.1; 98.1$ & $92.8\pm0.3; 97.7$ & $94.0\pm0.2; 97.8$ & $80.6\pm1.1; 93.6$ \\
      W2/A2 & $71.8\pm0.3; 82.6$ & $69.3\pm0.4; 80.6$ & $70.8\pm0.4; 81.6$ & $52.7\pm0.4; 61.3$\\
      \bottomrule
    \end{tabular}
    \label{tab:results}
    \vspace{-0.1in}
\end{table*}

\vspace{-0.1in}
\section{Discussion}
\label{sec:discussion}
\vspace{-0.05in}

As demonstrated, we can easily simulate quantization of CNNs to arbitrary precision. Furthermore, we could even model other scalar quantization methods. Thus, GHN-QAT becomes a powerful tool for quantization-aware design of efficient CNN architectures. The parameters predicted by GHN-QAT are remarkably robust and the QAT finetuning results (see Table~\ref{tab:results}) show a significant improvement over the full-precision float32 finetuning used in~\cite{ghnq_eiw}. This shows a clear benefit to adapting GHNs specifically to predict parameters for quantization-aware graphs. Additional possibilities/challenges of leveraging quantization-aware training, such as learned quantization thresholds or reducing QAT oscillations like in~\cite{nagel_qat_oscillations}, should be explored to further improve GHN-QAT, especially for "extreme" low bitwidths. It's possible that such improvements to QAT would make SimQuant more stable for 2-bit quantization. 

From GHN-QAT, we can see that introducing quantization into our GHN training allows for greater use of GHNs for quantization-specific neural network parameter prediction. Besides leveraging GHN-QAT for quantized versions of floating point operations, we should be able to encode quantization information such as bit-width and quantization scheme into the graphs. If used as a form of quantized accuracy prediction, GHN-QAT could greatly accelerate the process of searching for accurate, quantized CNNs. Additionally, GHN-QAT could be a useful weight initialization for quantized CNN training. The authors of~\cite{WhereShouldWeBegin} found noticeable differences in quantized accuracy of CNNs depending on their initialization. Thus, a quantization-aware weight initialization could be more robust than random initialization. If GHN-QAT-predicted parameters can be used as initialization for quantization-aware training rather than first training models to convergence in full float precision before additional QAT, then the training time of quantized models would be significantly reduced. Unfortunately, GHNs do not yet match the accuracy of iterative optimization and further improvements should be explored to bridge this gap. However, the aforementioned benefits could still greatly reduce the costs of designing quantized CNNs.

\bibliographystyle{IEEEtran}
\bibliography{main_ghnqat}

\end{document}